\documentclass[journal ]{new-aiaa}
\usepackage[utf8]{inputenc}
\usepackage{float}
\usepackage{placeins}
\usepackage{setspace}
\usepackage{textcomp}
\usepackage{graphicx}
\usepackage{doi}

\graphicspath{{figures/}}
\usepackage{amsmath}
\usepackage[version=4]{mhchem}
\usepackage{siunitx}
\usepackage{longtable,tabularx}
\usepackage[justification=justified,singlelinecheck=false]{caption}
\usepackage{etoolbox}
\patchcmd{\thebibliography}
  {\list}
  {\vspace{2.5pt}\list}
  {}{}
\hypersetup{hidelinks}
\begin{document}

\renewcommand{\thefootnote}{\arabic{footnote}}
\setcounter{footnote}{0}

\begin{center}

{\fontsize{20}{24}\selectfont\bfseries
A Systems Engineering Framework for\\
Vision-Language-Enabled UAV Triage and Disaster Response
\par}

\vspace{0.55cm}

{\fontsize{16}{18}\selectfont
Swapnil Saha\raisebox{0.6ex}{\fontsize{12}{18}\selectfont 1}
\par}

\vspace{2pt}

{\fontsize{12}{14}\selectfont\itshape
University of Arkansas, Fayetteville, Arkansas, 72701, USA
\par}

\vspace{0.35cm}

{\fontsize{16}{18}\selectfont
Bhuvan Rajanasiriyur Jagadeesha\raisebox{0.6ex}{\fontsize{12}{18}\selectfont 2}
~and~
Karishma Patnaik\raisebox{0.6ex}{\fontsize{12}{18}\selectfont 3}
\par}

\vspace{2pt}

{\fontsize{12}{14}\selectfont\itshape
University of Michigan-Dearborn, Dearborn, Michigan, 48128, USA
\par}

\vspace{0.35cm}

{\fontsize{16}{18}\selectfont
Neelakshi Majumdar\raisebox{0.6ex}{\fontsize{12}{18}\selectfont 4}
\par}

\vspace{2pt}

{\fontsize{12}{14}\selectfont\itshape
University of Arkansas, Fayetteville, Arkansas, 72701, USA
\par}

\end{center}

\footnotetext[1]{Graduate Research Assistant, Department of Mechanical Engineering, swapnils@uark.edu}
\footnotetext[2]{Graduate Research Assistant, Department of Electrical and Computer Engineering, bhuvanrj@umich.edu}
\footnotetext[3]{Assistant Professor, Department of Electrical and Computer Engineering, kpatnaik@umich.edu}
\footnotetext[4]{Assistant Professor, Department of Mechanical Engineering, neelm@uark.edu, AIAA Member}
\begin{abstract}
{\fontsize{10}{12}\selectfont
Recent advances in Vision Language Models (VLMs) have created new opportunities for disaster 
response, where responders must interpret large volumes of sensor data under critical time 
pressure. Current VLM applications in this domain include social media monitoring for situational 
awareness, generation of draft action plans, and translation of complex technical alerts into public
facing messages. While these efforts demonstrate the potential of VLMs to accelerate information 
flow, they remain largely limited to decision-support roles. In practice, such approaches can increase 
operator burden, as humans must still translate outputs into coordinated actions across teams and 
robotic assets. This study explores the viability of embedding VLMs as coordination agents within 
the human-UAV loop. The proposed architecture integrates natural language interaction, mission
level task coordination, software-in-the-loop implementation, and communication aligned with the 
Incident Command System (ICS). Rather than functioning solely as advisory tools, VLMs facilitate 
communication between human operators, mission control logic, and UAV task execution. The 
framework was developed using a Model-Based Systems Engineering (MBSE) approach, employing 
use case and block definition diagrams to represent system roles, internal structure, and component 
interactions. Three key components, the VLM Coordinator Agent, UAV Mission Control, and Task 
Allocator, were implemented within an integrated simulation and control environment. A preliminary 
human-factors evaluation with seven participants showed reduced perceived workload across mental 
demand, effort, and frustration, along with high ratings for AI trust and communication clarity. By 
integrating MBSE, software-in-the-loop testing, and human-factors evaluation, this work advances 
scalable human-autonomy teaming for high-stakes disaster response, with broader implications for 
aerospace autonomy and civil safety.
}

\end{abstract}

\section{  Introduction}
\vspace{0.2cm}
\lettrine{U}{nmanned} Aerial Vehicles (UAVs) are increasingly deployed for surveillance and monitoring, leveraging advanced Internet-of-Things (IoT) sensors and AI-driven traffic optimization systems that provide wide-area, real-time traffic data that reduces traffic congestion and monitors emissions \cite{moraga2025ai}. UAVs are part of the larger Advanced Air Mobility (AAM) ecosystem, which includes emerging aviation services enabled by electrification, autonomy, and on-demand transportation models \cite{phadke2025towards}. UAVs also serve as autonomous patrol agents in urban areas, detecting incidents such as street fights or fallen individuals and recommending appropriate responses without human supervision \cite{yuan2024patrol}. 
Beyond urban monitoring, they have proven critical in disaster response, providing rapid situational awareness, mapping, and reconnaissance in environments too dangerous or inaccessible for human responders. 

\begin{figure} [H]
    \centering

    \includegraphics[width=0.88\textwidth]{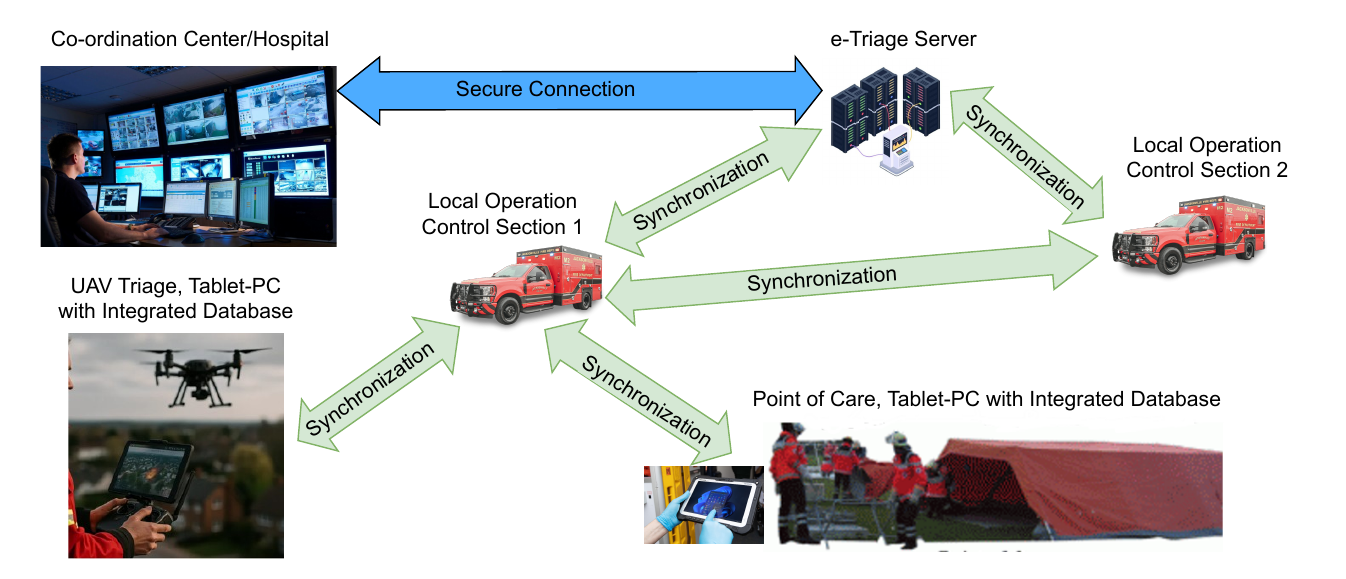}
    \captionsetup{justification=centering}
    \caption{Current e-Triage operational control architecture for individual rescue (adapted from \protect\cite{adler2011supported}). UAVs and field devices synchronize triage data with local control sections and a central server via secure links.}
    \label{fig:typical_MCI_flow}
\end{figure}

Commonly, digital platforms, such as kiosks or online portals, are employed to automate the initial assessment and prioritization of patients in a healthcare setting through electronic-triage (e-Triage) systems \cite{adler2011supported}. Figure \ref{fig:typical_MCI_flow} illustrates an overall e-Triage communication architecture, showing how UAV-based triage integrates into the emergency response network. Data from field tablets and UAV triage units synchronize with local operation control sections, which in turn connect to the central e-Triage server and coordination center through secure communication links. This structure enables rapid information exchange between on-site responders, medical tents, and hospitals, ensuring that situational awareness, triage data, and patient information are continuously updated and accessible across all levels of response. However, as fleet sizes and data streams increase, human operators face mounting cognitive demands to manage multiple UAVs, interpreting complex sensor inputs, and relaying information within the rigid structure of the Incident Command System (ICS). ICS is a hierarchical, role-driven framework governing information flow during emergency operations. Therefore, additional capabilities such as automated identification, and intelligent patrolling can be highly valuable in UAV triage and disaster response or recovery operations. 

Recent efforts have explored integrating language-based AI, in which Large Language Models (LLMs) enable semantic reasoning and task planning in robotic systems, including UAV operations. LLM is an advanced AI system that can understand and generate human language. It is combined with a vision encoder (Vision-Language Model or VLM) to make the system more powerful by allowing both text and visual data. Moreover, LLMs are used as robotic brains to improve performance in challenging scenarios, such as generating comprehensive action plans for manipulation tasks in complex environments \cite{wang2024large}. These artificial intelligence (AI) models can be beneficial when incorporated into UAV systems to support crisis response, as these environments are highly complex. 

While existing automation and AI aids primarily focus on flight control, image processing, or data summarization, the operator is still responsible for interpreting and coordinating outputs across teams. In this growing cognitive burden, Vision-Language Models offer a promising approach to reduce operator load, as they can interpret both visual and language-based information in a unified form, making their integration particularly relevant in this domain. With the integration of machine learning and VLMs, multi-UAV systems execute both synchronous and asynchronous missions, resulting in greater efficiency than traditional planning approaches \cite{cui2024tpml}. VLM-augmented UAVs can be used in mass casualty incidents, where the VLM component can work as a communication agent with patients, also aid classification of injury images \cite{mangel2024deployment}. VLMs also increase mission reliability by processing telemetry data and mission logs with contextual retrieval, resulting in high decision accuracy and low latency \cite{sezgin2025scenario}. Preliminary research shows that human-LLM synergy can enable teams with 126 UAVs to make operations smooth and energy-efficient for real-world disaster response scenarios \cite{sadik2025human}. Advanced technology, such as Spatial-Semantic Reasoning (the capability of LLMs to understand relationships among objects, locations, and meaning), with LLMs, can reduce mission duration by 36\% and path length by 50\% \cite{maletic2025spatial}. Together, these advancements are highly valuable for triage and crisis-response operations.

In real disaster management situations, despite these technological advancements, substantial human involvement is still required for mission planning, oversight, safety assurance, data validation, and related tasks, particularly in complex or dynamic environments where full autonomy remains a challenge. However, existing studies have not clearly shown how VLMs can be structured as coordination agents within a human-in-the-loop UAV triage workflow, or how such integration affects operator workload, trust, and communication clarity.
This paper investigates whether VLMs can be systematically integrated into the UAV-human communication loop to alleviate operator workload during disaster-response triage missions. To address this challenge, we developed a model-based systems engineering (MBSE) framework for integrating VLMs into UAV mission control, focusing on trust, workload reduction, and decision efficiency. MBSE diagrams were used to represent the framework of VLM and some key elements of VLM structure were implemented in software-in-the-loop simulation workflow. The implemented workflow was also tested through a preliminary human-factors experiment to examine how AI-assisted coordination affects perceived workload, trust, and communication clarity during UAV triage tasks.

\section{  Methodology}
\vspace{0.2cm}
\noindent\textbf{A. Model Based Systems Engineering (MBSE) Framework for VLM-Assisted UAV Triage}
\vspace{0.1cm}

This work adopts a model-based systems engineering (MBSE)-driven design approach to formalize the human-VLM-UAV coordination loop for triage missions. In this study, MBSE-driven design refers to the use of structured system models to represent system requirements, functional roles, component relationships, information flows, and mission behavior throughout the design process. The methodology consists of two progressive modeling layers using SysML visual representations (diagrams): (i) defining human interactions and mission roles (use case diagram); and (ii) representing the internal architectural structure of the system (block definition diagram).

 Figure 2 is a use case diagram that represents the high-level functional interactions between human and main system components involved in mission planning, constraint validation, task allocation, and triage reporting. Here, the rectangular shapes represent system domains, and the ovals represent the use cases or functions performed within each domain. In this scenario, the primary system authority is the VLM Control Interface which consists of a VLM Coordinator Agent and an Incident
Command System (ICS) interface. The VLM Coordinator Agent functions as the central decision-making module, while the ICS interface facilitates communication with the UAV and transmits triage data to VLM Coordinator Agent. At the mission initiation, the VLM Coordinator Agent activates the UAV and performs internal system checks, including battery status, navigation readiness, communication links, and sensor health. The Mission Constraint Validator and UAV Mission Control within the VLM Coordinator Agent assess available resources, validate safety limits, check airspace restrictions, evaluate flight routes, and generate the optimal action sequence before authorizing flight. During the mission, the UAV collects triage data and transmits it to the ICS interface, which organizes and structures the information according to incident command needs before forwarding it to the VLM Coordinator Agent for advanced reasoning. The VLM Coordinator Agent interprets the processed data into a human human-understandable format and relays it to the Operations Section Chief for final operational direction. Finally, the Operations Section Chief prioritizes decisions and issues the final command for the UAV to execute the mission.
\begin{figure} [H]
    \centering   \includegraphics[width=0.725\linewidth]{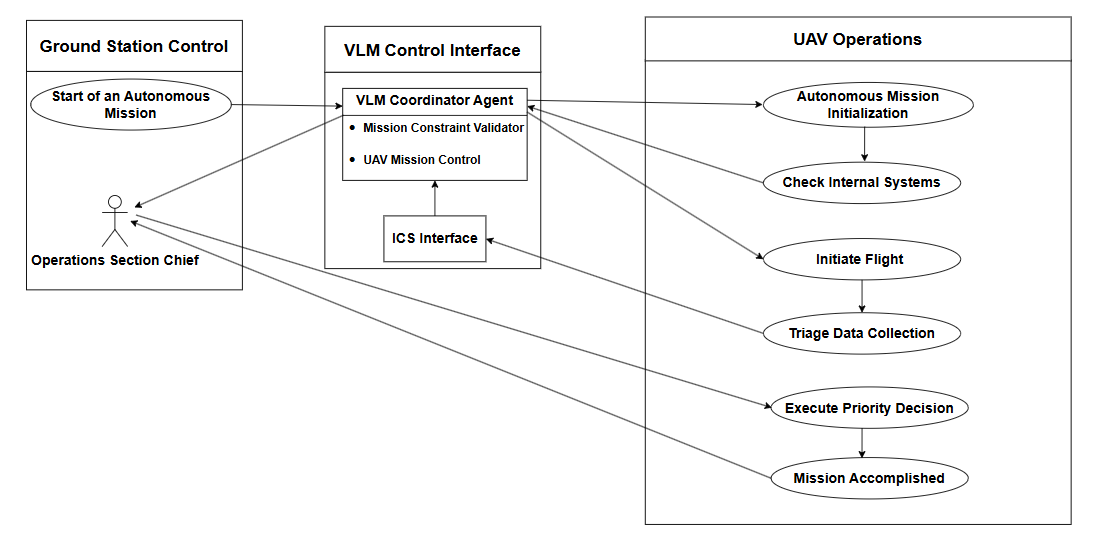}
    \captionsetup{justification=centering}
    \caption{VLM-assisted UAV coordination showing interaction among control interfaces and operations.}
    \label{fig:placeholder}
\end{figure}

 Fig. 3 represents the Block Definition Diagram (BDD) which illustrates the structural hierarchy of the system and defines the main system blocks, their properties, operations, and interrelationships. The BDD focuses on the VLM Control Interface, the central component shown in Fig. 2, serving as  the primary system authority. This diagram details how internal modules (VLM Coordination Agent, Mission Constraint Validator, Task Allocator, Mission Control, UAV Fleet, ICS Interface, and Audit \& Logging) are structured, decomposed, and interconnected within the overall architecture. This structural view enables traceability, modularity, and clear allocation of responsibilities prior to modeling dynamic execution flow. Each rectangular block represents a system element, with arrows indicating the direction of data interaction between components. Each element maintain its own properties and operations for system functionality. The operation syntax in BDD follows the format \emph{OperationName(parameterName : ParameterType) : ReturnType}. 
Here, the text before the colon is the input variable name, and the text after the colon is the type of information being passed into the operation. For example, the operation \emph{DeriveMissionPlan(nl : NL\_Command) : MissionPlan} in VLM Coordinator Agent means that the operation receives one input called \emph{nl}, where \emph{nl} stands for a natural-language command. A natural-language command is a human-readable instruction written in normal language, such as “inspect the highest-priority sector,” “return to base,” or “search the next assigned area.” The operation converts this instruction into a MissionPlan, which represents the structured mission output needed by the UAV system, such as the assigned task, target area, priority level, route direction, and safety constraints.

\begin{figure}[H]
  \centering
  \includegraphics[width=\textwidth,height=13cm,keepaspectratio]{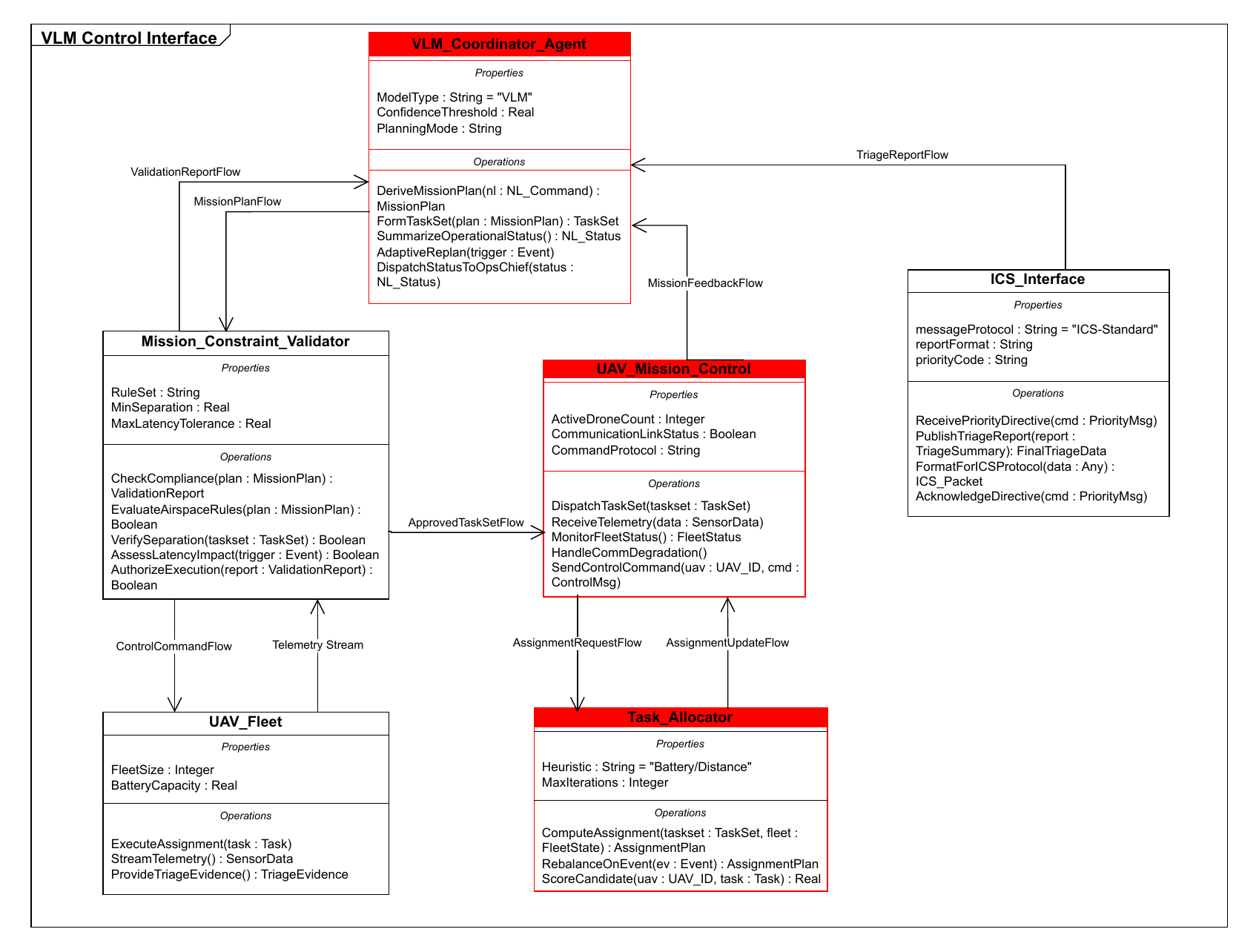} 
 
  \caption{Block Definition Diagram of the VLM Control Interface showing the main system blocks, their responsibilities, and data flows for mission planning, constraint validation, task allocation, UAV control, ICS communication, and telemetry feedback. The highlighted boxes represent the functions validated in this article in a simulation-in-the-loop framework.}
  \label{fig:usecase}
\end{figure}
In Fig. 3, the three highlighted BDD elements, the VLM Coordinator Agent, UAV Mission Control, and Task Allocator, were implemented in the software-in-the-loop workflow. The VLM Coordinator Agent was implemented to interpret mission-related inputs, summarize operational status, and support adaptive mission planning. The UAV Mission Control element was implemented to receive telemetry, monitor vehicle status, and send mission commands. The Task Allocator was implemented to assign mission tasks based on mission needs and vehicle status. Together, these implemented elements represent the core executable portion of the MBSE architecture, connecting VLM-based coordination, UAV control, and task-level decision support within the simulation environment. 

As shown in Figs. 2 and 3, the VLM is designed to manage most of the operational workload, while assisting human decisions. These diagrams support the objective of reducing human workload while retaining oversight at critical stages. Together, they establish a closed-loop coordination framework between human command and autonomous UAV operations, ensuring validated, adaptive, and mission-safe triage execution. 

\vspace{0.2cm}
\noindent\textbf{ B. Simulation-Based Implementation of the Proposed MBSE Framework}

\begin{figure}[htbp]
    \centering
    \includegraphics[width=1\linewidth]{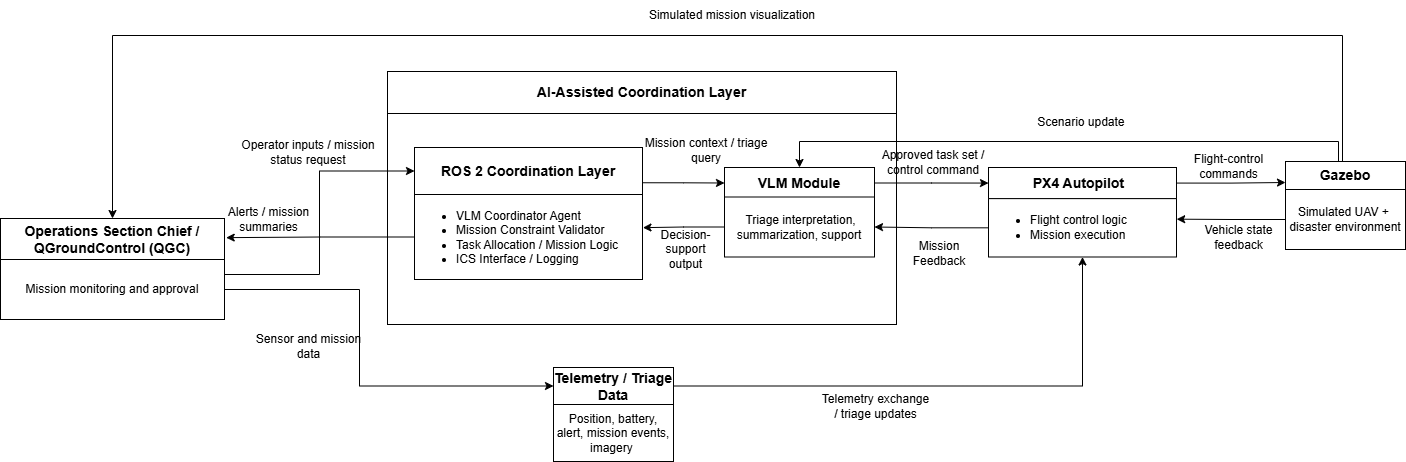}
    \caption{Complete control architecture for the proposed MBSE framework using ROS 2, PX4, Gazebo, QGroundControl, and a VLM-based coordination layer.}
    \label{fig:simulation_workflow}
\end{figure}

\vspace{0.1cm}
To demonstrate how the proposed MBSE architecture can be translated into an executable environment, the framework was implemented through a software-in-the-loop (SITL) workflow using using ROS 2, Gazebo, PX4, and QGroundControl (QGC), as shown in Fig.~\ref{fig:simulation_workflow}. ROS 2 was used as the middleware layer for communication between the UAV simulation, mission logic, and operator-facing interface. ROS provides open-source software libraries and tools for developing robotic systems, making it suitable for connecting distributed components in the proposed UAV triage workflow~\cite{ros,ros2docs}. Gazebo was used to provide virtual disaster-response environment and simulated UAV platform. As an open-source robotics simulator, Gazebo supports physics-based simulation, rendering, and sensor modeling, which enables the UAV mission to be tested in a safe and repeatable environment~\cite{gazebo}. PX4 was used as the flight-control and autopilot layer. It enables vehicle state updates, mission execution, and autonomous behavior within the simulated UAV system~\cite{px4}. QGroundControl was used as the ground-station interface through which the human operator can monitor vehicle status, mission progress, and operational alerts during the simulation~\cite{qgc}.
This implementation is consistent with the system roles and interactions defined earlier in the use case diagram, the structural decomposition shown in the Block Definition Diagram (BDD). In this way, the MBSE model is not limited to a conceptual design, but is connected to a practical software and simulation environment for UAV triage operations.

\FloatBarrier

\vspace{0.25cm}
Figure~\ref{fig:simulation_workflow} illustrates the software-in-the-loop workflow used to validate the proposed MBSE architecture. The operator-facing QGroundControl interface represents the human supervision layer, consistent with the Operations Section Chief role in the MBSE framework. The AI-assisted coordination layer contains the ROS 2 Coordination Layer and the VLM Module. ROS 2 manages communication among mission logic, telemetry updates, task allocation, ICS-related functions, and logging, while the VLM Module supports mission-context interpretation, triage summarization, and decision-support output.

PX4 Autopilot represents the UAV mission-control function by handling flight-control logic, mission execution, and vehicle-state feedback. Gazebo provides the simulated UAV and disaster-response environment. The Gazebo-PX4 connection enables software-in-the-loop testing, where PX4 sends flight-control commands to the simulated UAV, and Gazebo returns vehicle-state feedback. QGroundControl displays mission status and receives operator inputs, while telemetry and triage data, including position, battery state, alerts, mission events, and imagery, are exchanged with the coordination layer.

Within this workflow, mission context and triage queries are passed from ROS 2 to the VLM Module, and the VLM returns decision-support outputs to ROS 2. ROS 2 then sends approved task sets or control commands to PX4, which executes them in the Gazebo environment. This structure keeps the VLM as a decision-support component rather than a direct flight controller, while preserving the human-in-the-loop role through QGroundControl. Overall, the workflow shows how the conceptual MBSE elements are translated into interacting software, simulation, and operator-facing components for UAV triage evaluation.

\vspace{0.2cm}
\noindent\textbf{C. Human-Factors Workload Evaluation}
\vspace{0.1cm}

A preliminary human-factors evaluation was conducted with seven participants to assess interaction with the UAV triage simulation environment. The study was conducted under an Institutional Review Board (IRB)-approved protocol, with participants aged 18 years of age or older. The experiment used the ROS 2, Gazebo, PX4, and QGroundControl-based workflow. The participant task was to monitor a simulated UAV triage mission, track vehicle status and mission updates through QGroundControl and terminal interfaces, respond to task-related prompts, and compare the baseline interface with the AI-assisted coordination support. The interfaces displayed mission status, vehicle behavior, system information, and operational prompts throughout the task. Each session consisted of a brief 5-minute demonstration/training phase to familiarize participants with the task environment and controls, followed by an approximately 5-minute actual task phase. Only responses from the actual task phase were used for analysis.

\begin{figure}[H]
    \centering
    \includegraphics[width=0.49\linewidth]{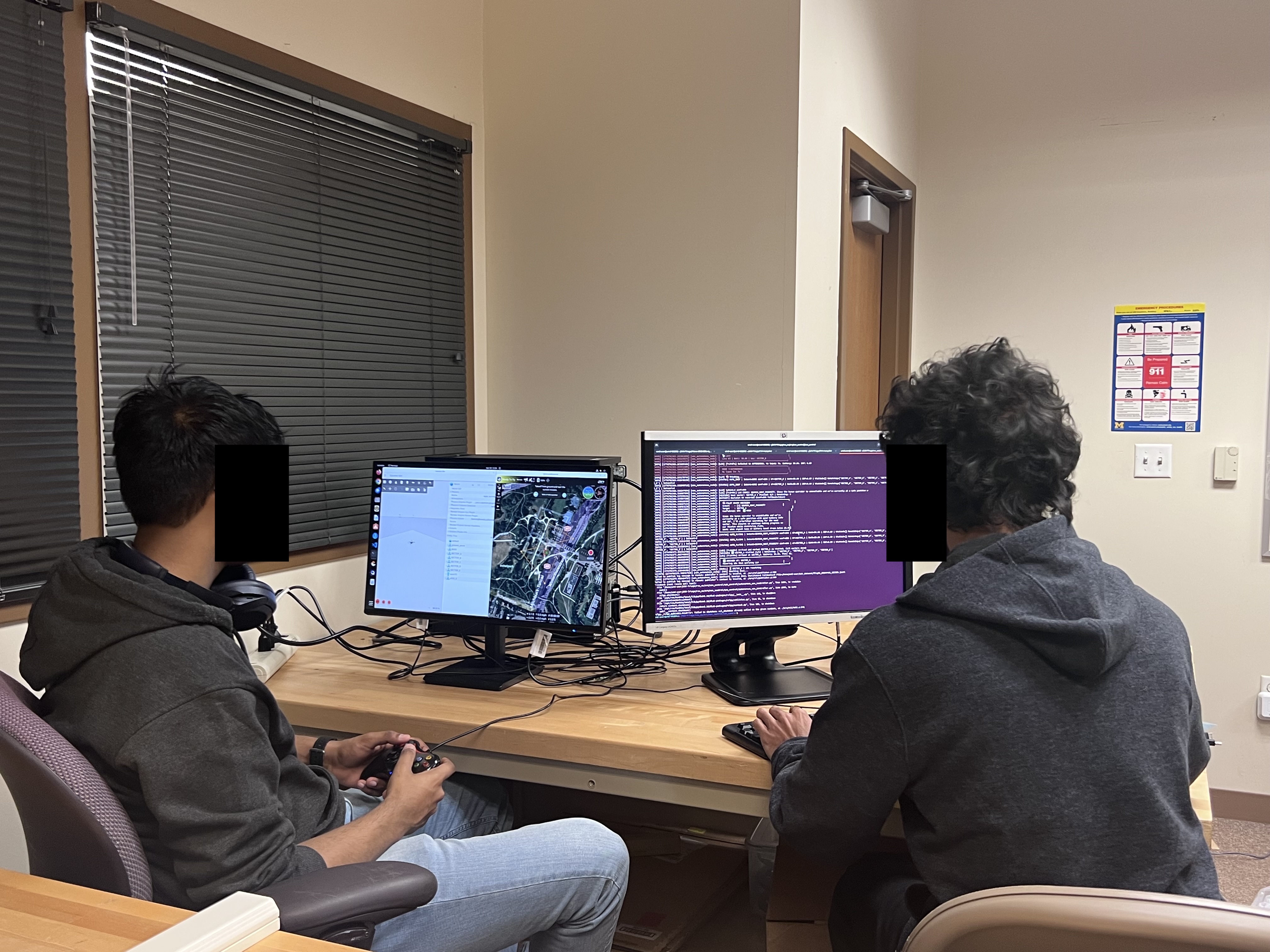}
    \caption{Participant interaction with the proposed and developed UAV triage simulation environment with VLM-in-the-loop.}
    \label{fig:human_experiment_setup}
\end{figure}
\FloatBarrier
As shown in Fig. 5, the participant interacted with the simulated UAV mission while the monitoring and command interfaces were displayed on the workstation. This setup represents the human-in-the-loop part of the proposed framework, where the operator observes mission progress, monitors system information, and responds to task-related events. 

Each participant completed two conditions: (1) a baseline condition and (2) an AI-assisted condition. In the baseline condition, participants monitored the UAV mission using the standard simulation interfaces, such as QGroundControl and terminal outputs, without additional AI guidance. In the AI-assisted condition, participants used the same simulation environment, and also received AI-based support in the form of mission interpretation, task-related guidance, and coordination prompts. This comparison was used to examine whether AI assistance could reduce perceived workload and improve user confidence during UAV triage tasks.

After completing the tasks, participants answered a short questionnaire using a 1 to 5 rating scale. For workload-related questions, 1 indicated very low and 5 indicated very high. For agreement-based questions, 1 indicated strongly disagree and 5 indicated strongly agree. Participants rated two conditions: the baseline condition and the AI-assisted condition. The participants rated the baseline condition using three workload-related factors: (1) mental demand, (2) effort, and (3) frustration. The same three factors were also rated for the AI-assisted condition. These questions were included to compare how participants perceived the levels of difficulty, effort, and frustration in the two conditions. 

The questionnaire also included targeted items for the AI-assisted condition. AI trust was assessed through two questions: (1) whether the AI assistant’s support felt reliable; and (2) whether the AI assistant’s behavior was predictable. Communication clarity was assessed through two items examining whether the AI assistant’s prompts made coordination more efficient and whether the AI-generated language was clear and well-structured. Participants also reported their overall preference, which condition felt more mentally demanding, and whether they would choose the AI-assisted mode for a similar task in the future. Together, these elements establish a basis for evaluating the viability of AI-assisted communication in operational human-autonomy teaming contexts.

All workload-related responses were summarized using a NASA Task Load Index-(TLX)-style chart. The NASA-TLX is a widely used subjective workload assessment method that evaluates how demanding a task feels to a participant \cite{hart2006nasa}. In this study, three NASA-TLX subscales were selected to capture key aspects of workload: mental demand, effort, and frustration. Mental demand represents the amount of mental and perceptual activity required during the task, such as thinking, deciding, remembering, looking, and searching. Effort represents how hard the participant had to work, both mentally and physically, to achieve the required level of performance. Frustration represents how insecure, discouraged, irritated, stressed, or annoyed the participant felt during the task \cite{HART1988139}. 
These three subscales were most relevant for a stationary, communication-focused task where physical and temporal demands were constant across conditions. 
The NASA-TLX-style chart was used to compare these workload ratings between the baseline and AI-assisted UAV triage conditions.
\section{  Results}
\vspace{0.1cm}
We analyzed responses from a total of seven participants regarding their comfort level with mouse and keyboard controls. One reported low comfort, another reported neutral comfort, and the remaining five participants reported being comfortable. All participants had prior experience with games requiring rapid keyboard input (e.g., WASD movement keys and hotkeys).
The results are organized into three parts: (1) mean NASA-TLX-style workload comparison; (2) workload distribution using box plots; (3) trust and communication in the AI-assisted condition.

Figure 6 presents an overall comparison of perceived workload between the baseline and AI-assisted conditions across three subscales: Mental Demand, Effort, and Frustration. The radial scale ranges from 1 (very low) to 5 (very high), with lower values indicating lower perceived workload. The radar chart shows that the AI-assisted condition produced lower scores across all three workload subscales compared with the baseline condition. Mental Demand decreased from 2.29 in the baseline condition to 1.43 in the AI-assisted condition. Effort decreased from 2.43 to 1.43, and frustration from 2.86 to 1.86.  These results indicate reduced perceived workload under the AI-assisted condition compared to the baseline condition. Among the three subscales, frustration exhibited the highest baseline value, suggesting it was the most noticeable burden in the non-AI workflow. The consistent reduction across all three subscales further suggests that the AI support helped reduce overall operator workload during the simulated UAV triage task.
\begin{figure}[H]
    \centering
    \includegraphics[width=0.75\textwidth]{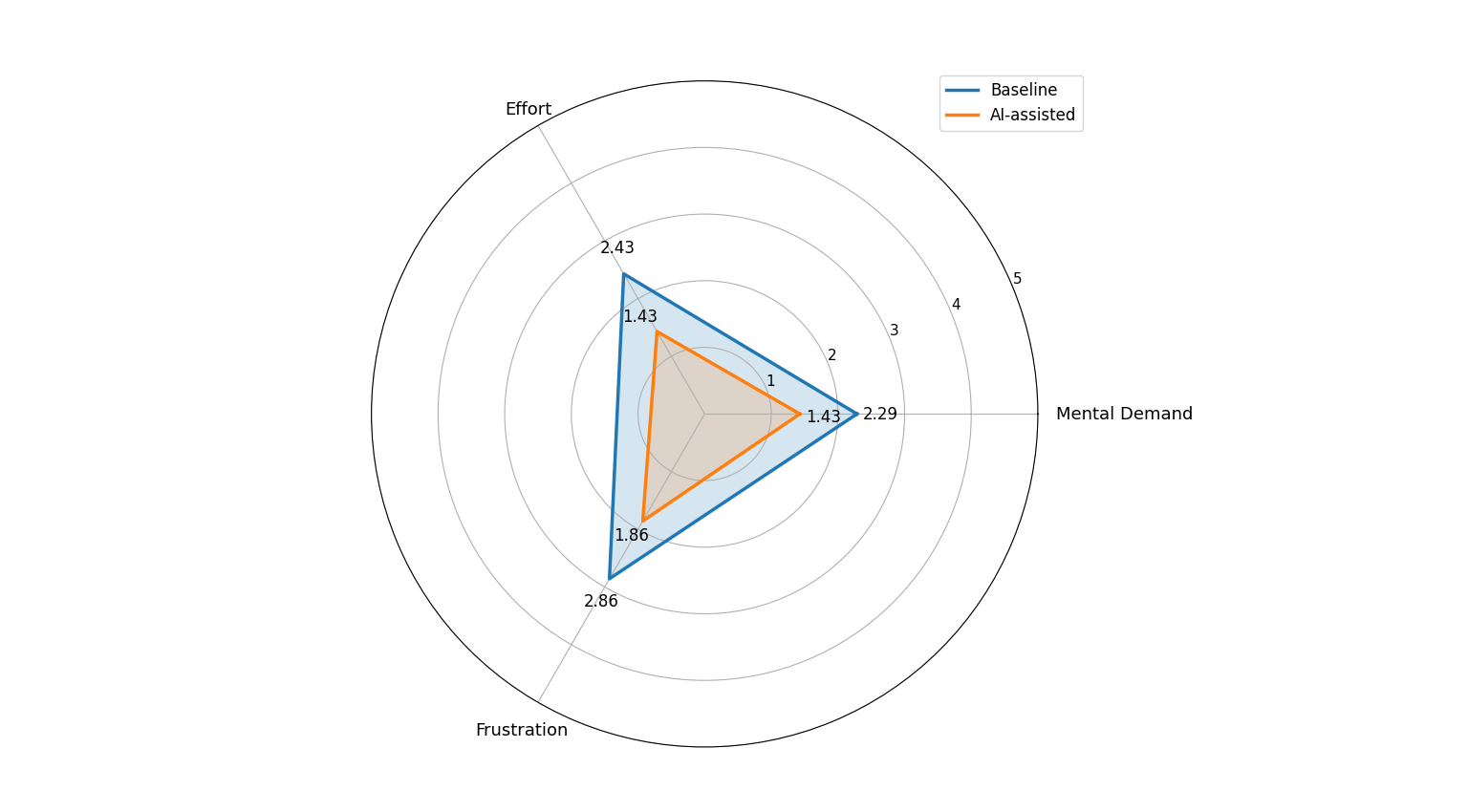}
    \caption{NASA-TLX-style workload comparison between baseline and AI-assisted UAV triage conditions.}
    \label{fig:nasa_tlx_radar}
\end{figure}

Figure 7 shows the participant-level distribution of workload ratings using box plots. For these workload factors, lower values indicate lower perceived workload. Overall, the AI-assisted condition had lower mean and median workload ratings across all three factors. The baseline condition produced more high-end responses and outliers, especially for effort and frustration. For Mental Demand, the median decreased from 3 in the baseline condition to 1 in the AI-assisted condition. The mean value also shifted downward, showing that participants generally experienced lower cognitive demand when AI support was available. The AI-assisted responses were concentrated near the lower end of the scale, indicating more consistent low workload ratings.

For Effort, the median decreased from 2 in the baseline condition to 1 in the AI-assisted condition. This suggests that participants felt the task required less manual and mental effort with AI assistance. One outlier near 4 appears in the baseline effort ratings, meaning one participant reported noticeably higher effort without AI support.

\begin{figure}[H]
    \centering
\includegraphics[width=0.7\linewidth]{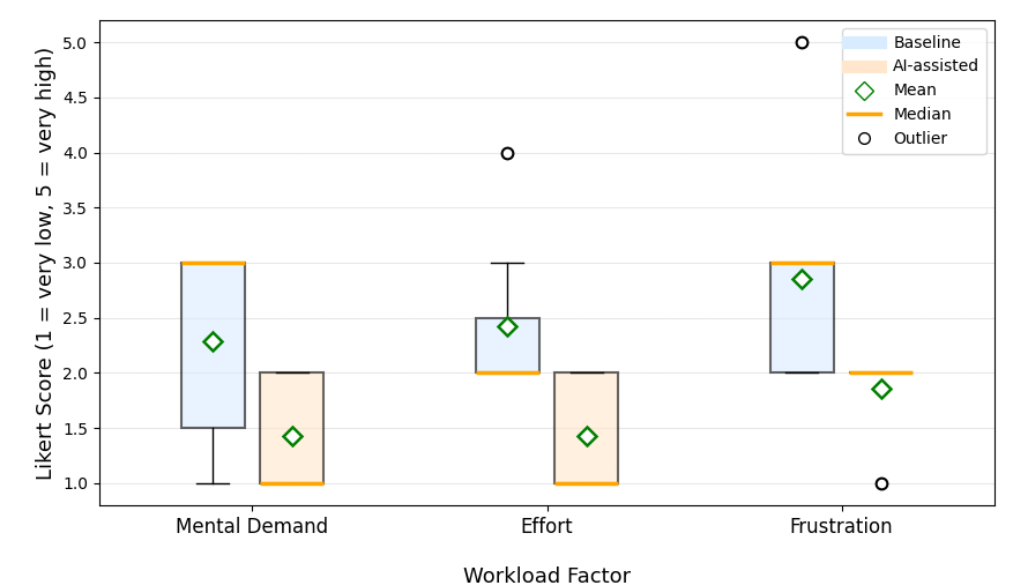}
    \caption{Distribution of participant workload ratings for mental demand, effort, and frustration in baseline and AI-assisted UAV triage conditions. The box represents the middle 50\% of responses, the orange line represents the median, the green diamond represents the mean, the whiskers show the main spread of the data, and the circle markers indicate outliers.}
    \label{fig:workload_boxplot}
\end{figure}
For Frustration, the median decreased from 3 in the baseline condition to 2 in the AI-assisted condition. The baseline condition also includes a high outlier near 5, indicating that one participant experienced very high frustration in the non-AI task. In contrast, the AI-assisted frustration ratings were mostly concentrated around lower values, , indicating reduced perceived burden. These results suggest that AI assistance reduced perceived workload and improved the manageability of the UAV triage task for participants.

\vspace{0.1cm}
Figure 8 shows the participant ratings for Trust and Communication Clarity in the AI-assisted condition. These two factors were measured only for the AI-assisted mode because the baseline condition did not include an AI assistant. The scale ranges from 1 (strongly disagree) to 5 (strongly agree). Therefore, higher values indicate better trust and clearer communication. These ratings provide additional insight into how participants perceived the AI assistant beyond workload reduction. They also help evaluate whether the AI-generated information was understandable, reliable, and useful during the simulated UAV triage task. Higher ratings in these categories suggest that the AI-assisted system supported operator confidence and improved the clarity of decision-support communication.
\begin{figure}[H]
    \centering
    \includegraphics[width=0.65\linewidth]{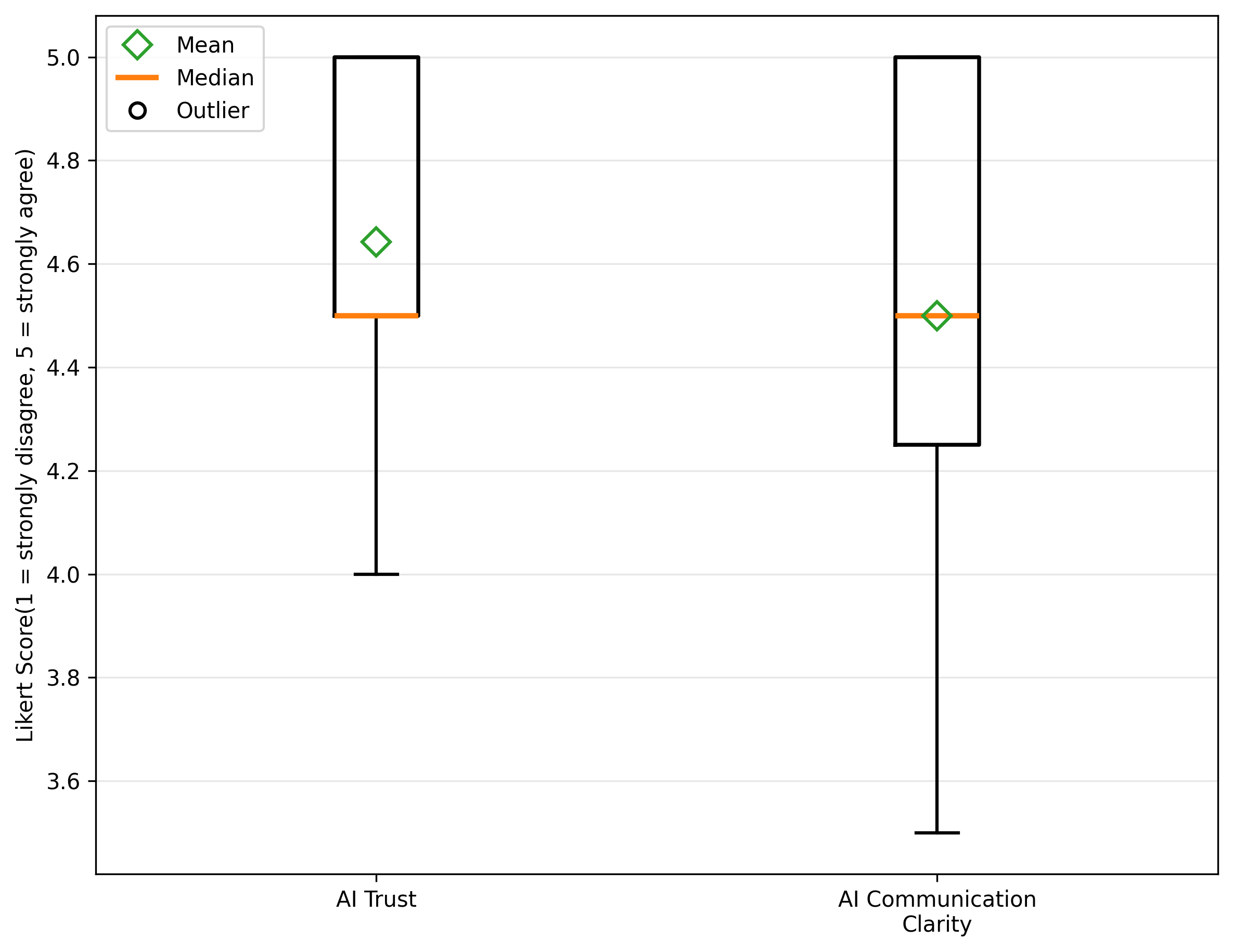}
    \caption{Participant ratings of AI trust and communication clarity in the AI-assisted UAV triage condition.}
    \label{fig:ai_trust_comm}
\end{figure}

Overall, participants gave high ratings for both trust and communication clarity. For AI Trust, the mean score was 4.64, and the median was 4.50. The responses were concentrated between 4.0 and 5.0, indicating that participants generally found the AI assistant reliable and predictable throughout the UAV triage task.

For AI Communication Clarity, the mean and median scores were both 4.50, with most ratings between 4.50 and 5.00, and a minimum value of 3.50. These results indicate that participants generally found the AI prompts clear, well-structured, and useful for coordination. However, the lower bound suggests that one participant perceived the communication clarity as moderately strong rather than very strong.

Together, these findings support the workload findings by showing that the AI-assisted workflow not only reduced perceived workload but also provided support in a way that participants found understandable and dependable. Since UAV triage tasks  
require quick decisions under time pressure, high trust and clear communication are critical for maintaining effective human-in-the-loop operation.

\section{  Conclusion}
\vspace{0.1cm}
This paper presented an MBSE framework for a VLM-assisted UAV triage system in disaster-response operations, connecting conceptual system architecture from MBSE to an executable software-in-the-loop workflow. The key contributions are: (1) an MBSE-based model of human-VLM-UAV coordination capturing system roles, interactions, and operational flows; (2) a software-in-the-loop implementation linking the VLM Coordinator Agent, UAV Mission Control, and Task Allocator in an executable UAV triage environment; and (3) a preliminary human-factors evaluation demonstrating reduced perceived workload and high trust and communication clarity under the AI-assisted condition. Overall, the findings suggest that VLM-assisted coordination can improve the operational effectiveness of human-UAV triage tasks while keeping the human operator in the decision loop.

Given the preliminary nature of this evaluation with seven participants, the results are reported descriptively and should be interpreted with caution. Future work should extend the current software-in-the-loop workflow to a real UAV platform to validate the VLM-assisted coordination system in physical flight operations. The remaining elements of the Block Definition Diagram, including the Mission Constraint Validator, ICS Interface, and UAV Fleet modules, should be completed to fully represent the MBSE architecture in the executable workflow. Future studies should increase the participant sample size to strengthen the human-factors evaluation of workload, trust, and communication clarity. More realistic disaster-response scenarios, such as multiple victims, low-battery alerts, communication delays, and changing mission priorities, should also be included. Finally, the framework should be extended to multi-UAV configurations to evaluate scalability in larger emergency-response missions.

\section*{Acknowledgments}
\vspace{0.1cm}
This work was supported by the Department of Mechanical Engineering at the University of Arkansas and the Department of Electrical and Computer Engineering at the University of Michigan-Dearborn.
\vspace{-0.45cm}
\begin{singlespace}
\setlength{\bibsep}{0pt}
\bibliography{references}

\end{singlespace}
\par\vspace{100pt}
\end{document}